\documentclass[11pt]{article}

\usepackage[final]{acl}

\usepackage{times}
\usepackage{latexsym}

\usepackage[T1]{fontenc}

\usepackage[utf8]{inputenc}

\usepackage{microtype}
\usepackage{inconsolata}
\usepackage{graphicx}
\usepackage{linguex}
\usepackage{amsmath}
\usepackage{amssymb}

\title{Energy-Based Transformers as Predictors of Reading Difficulty}

\author{Jakub Dotla\v{c}il \\
  Utrecht University \\
  \texttt{j.dotlacil@uu.nl} \\\And
  Ece Takmaz \\
  Utrecht University \\
  \texttt{e.k.takmaz@uu.nl} \\}

\begin{document}
\maketitle

\begin{abstract}
Transformer language models have become established tools for modeling human sentence processing, with measures such as surprisal and attention entropy serving as effective predictors of reading difficulty that together capture complementary aspects of processing load. Here, we explore a related class of transformer models: energy-based transformers, which provide a principled formal link to associative memory models, bringing processing research into direct contact with the broader literature on Hopfield networks and dense associative memory. To our knowledge, this is the first exploration of an energy-based transformer measure in computational psycholinguistics. Across reading-time corpora (Natural Stories, UCL eye-tracking, UCL self-paced reading), the energy measure is a robust predictor of reading times, providing significant fit beyond surprisal and entropy in all three. In a controlled experiment on relative clause processing, energy at a single layer captures the well-known object/subject asymmetry. We find evidence that it subsumes effects attributable to both attention entropy and surprisal, suggesting that energy may serve as a single unified predictor where multiple complementary measures have previously been required.
\end{abstract}

\section{Introduction}

Transformers are dominant models not only in natural language processing and machine learning, but also in computational psycholinguistics as models of human sentence processing. The link to human sentence processing literature has been established in two main ways. First, the \textbf{surprisal theory} of processing \citep{hale2001,levy2008} made an argument that the surprisal (i.e., the log of the inverse probability) of the next token can be used to predict neurobehavioral data. Surprisal from transformer models can be easily collected and used for this prediction. Second, it has been noticed that the \textbf{attention mechanism in transformers} conceptually resembles memory processing accounts like \textbf{cue-based memory} \citep{lewis2005}. In support of that, some summarizing measures of attention, like \textbf{attention entropy}, have been successfully used as a predictor of memory interference in human processing \citep{ryu2025}.

There are currently at least two challenges for the link between transformers and human sentence processing. First, we do not have one overarching measure that would fit processing data such as reading times. For instance, as we will see in Experiment~1 below, attention entropy and surprisal are both useful and needed for modeling reading-time data. Finding a single measure that on its own captures predictions of both of these measures would substantially strengthen the link between the models and sentence processing. Second, while the conceptual link between memory-based models and attention is intuitively appealing, we lack a clearer link between transformers and human memory (see also \citealt{oh2026}). Having such a link would allow us to investigate the connection between memory and processing in transformers beyond the appeal to just our conceptual understanding and intuitions.

In a different tradition, cognitive scientists and machine learning researchers developed and implemented models of \textbf{associative memory} called Hopfield Networks \citep{hopfield1982}. There are two recent developments that should make computational psycholinguists consider such models. First, while classical Hopfield networks had very limited storage capacity, this issue has been overcome in theoretical developments that led to so-called modern Hopfield networks or dense associative memory \citep{krotov2016}. These changes, we believe, make the memory capacity of these models more realistically human-like, making them more suitable for studying cognitive tasks that require rich memory representations, with human sentence processing being a prime example. Second, very recent theoretical investigations showed that dense associative memory is closely related to transformer attention \citep{ramsauer2020} and some versions of \textbf{energy-based transformers} can in fact be seen as particular implementations of the dense associative memory theoretical framework \citep{hoover2023, krotov2025}.

The last link is the starting point of this work. We explore one particular implementation of energy-based transformers, namely \textbf{energy GPT} (\textbf{NRGPT}, \citealt{nrgpt}) and study how it can be linked to predicting processing. We show that the scalar function that is at the core of NRGPT, energy, can naturally be connected to reading-time data and we provide evidence that the energy of NRGPT captures at least some processing difficulties that currently require the combination of surprisal and attention entropy, thus replacing a mixture of measures with a single, straightforward link between transformers and processing. Finally, we fit NRGPT to a reading experiment on relative clauses and to several reading-time corpora to show that \textbf{the energy measure can capture reading patterns beyond the surprisal and attention-entropy measures alone}.

\section{Related work}
\label{sec:related}

Using transformers as models of human sentence processing has been popular in recent years in psycholinguistics, with the main focus on surprisal \citep{hao2020, merkx2021, hahn2022, hoover2023b, wilcox2023}. Beyond surprisal, a number of other measures have been considered for reading-time data, often based on the attention mechanism, such as attention entropy \citep{ryu2021, oh2022, ryu2025}. A recurring complication for the surprisal-based approach is that, with increasing model size, transformers actually get \emph{worse} at predicting reading-time data \citep{oh2023, kuribayashi-etal-2024-psychometric, michaelov2026}.

This scaling behavior has shaped which models are used. GPT-2, and in particular GPT-2-small, has become the standard choice for predicting human sentence processing, since it was found to sit in the best spot: it fits the data better than other, smaller models \citep{hao2020}, while larger models show a decrease in fit \citep{oh2022surprisal,oh2023}. Further, the decrease in fit with the growth of model size led to investigations into whether memory is at least partially responsible here. There are various attempts to inhibit the memory of transformers to make their next-word predictions more human-like \citep{hahn2022, timkey-linzen2023, lesci2024, thamma2026humanlike, michaelov2026}. In contrast to that research, we use models that were inspired by the properties of human associative memory from the start, and study how versions of them adapted to the transformer architecture can be useful in predicting human processing data. This potentially gives us a way to link, in a principled way, modern research on associative memory and sentence processing. If this research line is successful, we should in the future be able to translate findings on memory, formalized in the models of dense associative memory, into research on processing. Here, we establish the first step, by showing that the energy-based models originating from dense associative memory research can be linked to and predict human reading-time data. \textbf{To the best of our knowledge, this article is the first investigation of energy-based transformers in computational psycholinguistics for the purpose of fitting them to human sentence processing.}

\section{Theoretical background: NRGPT}

\begin{figure}[h]
  \includegraphics[width=\linewidth]{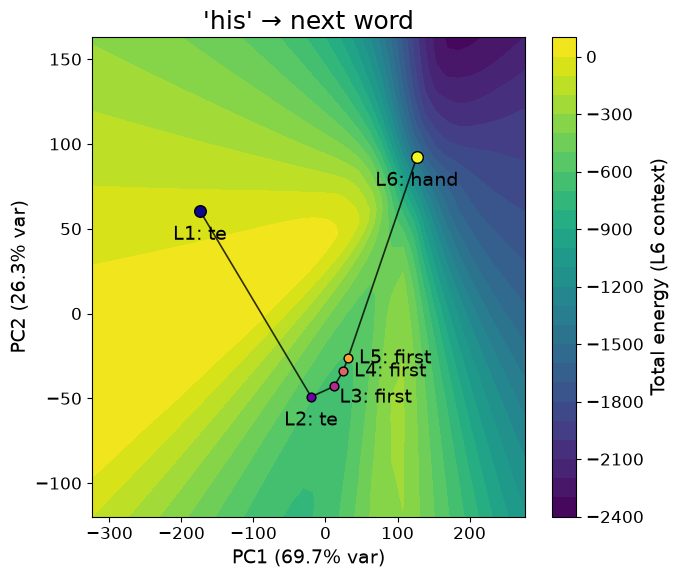}
  \caption{Descent through the energy landscape. L1\ldots L6 - layer iterations at the word \textit{his} in the sentence prefix \textit{He unlocked the door with his \ldots} The predicted words show the projection of LM head per layer. \label{fig_ex}}
\end{figure}

NRGPT \citep{nrgpt} is a recent reformulation of the GPT architecture that recasts a transformer's forward pass as gradient descent on a scalar energy function defined over token states. At each step, the token representations are directed downhill along the energy landscape, so running the model can be understood as moving toward a low-energy configuration for the input.  Figure~\ref{fig_ex} illustrates this: as layer iterations proceed, the energy decreases and the predicted next token becomes progressively more plausible.

Architecturally, NRGPT stays close to a standard GPT: it operates on the same input embeddings, uses multi-head causal attention, has a feedforward sublayer, and produces next-token logits at the output. There are, however, some structural changes. We discuss the two significant differences from conventional transformers: NRGPT uses a parallel structure, and it applies the transformer block recurrently. In the discussion below, $x\in\mathbb{R}^{D\times N}$ is the input sequence with $N$ tokens and $D$ dimensions and $x^{(t)}$ represents the input sequence at the layer $t$ (for standard GPT) or the layer iteration $t$ (for NRGPT).

\paragraph{Parallel-style transformer.} A conventional transformer is sequential: a layer-normed input sequence is fed into an attention layer (AT); the attention output is added back to the input, re-normalized (LN), and fed into a feed-forward layer (FF), whose output is added in turn:

\begin{equation}
\begin{aligned}
  x^{(t+1)} = {}& x^{(t)} + \text{AT}(g^{(t)}) \\
  &+ \text{FF}(\text{LN}(x^{(t)} + \text{AT}(g^{(t)}))),
\end{aligned}
\end{equation}
where $g^{(t)} = \text{LN}(x^{(t)})$ is the (layer-)normalized hidden state. By contrast, NRGPT is a parallel-style transformer, i.e., the attention and feed-forward layer apply in parallel:

\begin{equation}
x^{(t+1)} = x^{(t)} + \text{AT}(g^{(t)}) + \text{FF}(g^{(t)})
\end{equation}

The adherence to the parallel structure is strictly speaking not necessary for energy-based transformers, but this design choice makes it easier to implement next-token prediction as a descent through the energy landscape; see below.

\paragraph{Recurrence.} A standard transformer applies a stack of distinct layers in a single forward pass, whereas NRGPT applies the transformer block recurrently for $k$ iterations. Each iteration corresponds to one descent step on an energy landscape, which we introduce next.

\paragraph{Energy.} NRGPT associates a scalar energy $E_A$ with each token position $A$, decomposed into the attention energy $E^{\textit{AT}}_A$ and the feedforward energy $E^{\textit{FF}}_A$,
\begin{equation}
E_A(g) = E^{\textit{AT}}_A(g) + E^{\textit{FF}}_A(g).
\end{equation}
Intuitively, the energy can be seen as a processing effort. It is low when the token at position $A$ is easy to integrate and fits the context and the past knowledge, and high when it is difficult to integrate.

\paragraph{Inference as gradient descent.} The generation of tokens in NRGPT proceeds by minimizing each per-token energy $E_A$ with respect to its own token state. The update rule that replaces a conventional transformer layer is
\begin{equation}
  x_A^{(t+1)} = x_A^{(t)} - \eta^{(t)} \, \frac{\partial E_A}{\partial g_A^{(t)}},
\label{eq:nrgpt-update}
\end{equation}
where $\eta^{(t)} \in \mathbb{R}^{D\times D}$ is an (optionally learnable) inference-rate matrix and $g_A^{(t)}$ is the normalized state at position $A$. Applying \eqref{eq:nrgpt-update} for $k$ steps produces a trajectory $x_A^{(0)}, x_A^{(1)}, \dots, x_A^{(k)}$ on token $A$'s energy landscape. The final state $x_A^{(k)}$ is then mapped to next-token logits via the unembedding language model head in the usual way.

\paragraph{Attention and feedforward components.} The two energy components, $E^{\textit{AT}}_A(g)$  and  $E^{\textit{FF}}_A(g)$ are built so that differentiating them recovers the ordinary attention and feedforward operations of a transformer. Intuitively, one NRGPT iteration behaves much like one parallel-style transformer layer (with weights tied across iterations), except that the layer is now obtained by descending an energy landscape rather than being stipulated directly (see also Figure~\ref{fig_ex} for a concrete visualization in a 2D PCA reduction, with a label that would be obtained if the language model head was applied). We note that the first energy component, $E^{\textit{AT}}_A(g)$, is a negated log-sum-exp over the attention scores and corresponds to dense associative memory with a log-sum-exp interaction function (for details, see \citealt{ramsauer2020}). The second energy component, $E^{\textit{FF}}_A(g)$, is dense associative memory with another non-linear interaction function. Further details on these components and on how their gradients reproduce the standard attention and feedforward sublayers are discussed in \cite{nrgpt} and are briefly summarized here in Appendix~\ref{app:components}.

\paragraph{Why energy is informative.} Two properties of this construction matter for our purposes. First, energy is a single scalar summary of the network's state at each token and each iteration, well defined for any model trained with the NRGPT objective. Second, NRGPT can be shown to be asymptotically stable: under mild conditions, the energy of each token decreases along the trajectory and converges to a fixed value once the preceding tokens have settled \citep{nrgpt}. The trained model used in practice is not constrained to satisfy these mild conditions, so the descent is likely and often observed, but not required, that is, individual iterations need not always lower $E_A$, but they often do so (e.g., from layer 1 onward in our relative clause experiments, see below). The energy profile across iterations $t = 1, \dots, k$ therefore can be seen as a scalar summary of token's processing trajectory, providing a dynamical view of how the model integrates each token into the ongoing representation. In the experiments below, we use NRGPT with $k = 6$ iterations and treat the energy at intermediate iterations as a candidate predictor of human reading times.

\paragraph{Energy and transition probability.} A key interpretive point connects NRGPT's energy to the surprisal literature. Within the energy-based modeling framework, an energy function is intended by convention to play the role of a negative log-likelihood, so that low-energy configurations correspond to high-probability ones. \citet{nrgpt} appeal to this convention for NRGPT: at each position $A$, the state $g_A$ starts from the previous token's embedding and is moved by the dynamics in \eqref{eq:nrgpt-update} to a state of lower energy, and the per-token energy $E_A$ at that position can be read as scoring the (negative log) transition probability for the upcoming token given its preceding context. The interpretation is most cleanly motivated in the simplified regime where $\eta = I$. This does not hold in our setup, where $\eta$ is a learnable matrix. For the model below, we thus treat the identification as an interpretive frame rather than a strict equality. With that caveat, energy is the natural analog of surprisal: surprisal scores the lexical identity of the upcoming word under a discrete softmax readout, while energy scores the cost of the continuous-state integration trajectory that produces it. Both quantities are tied to the same underlying conditional dynamics, but they need not coincide. A predictable word reached via a costly integration can have low surprisal and high energy, and a surprising word that fits readily into the parse can have high surprisal and relatively low energy. This is the basis for our hypothesis that energy captures integration-driven difficulty that surprisal misses, while still tracking the predictability-driven component that surprisal is sensitive to.

\paragraph{Energy and attention entropy.} The attention energy is also tied to attention entropy, the measure that \citet{ryu2025} use as a predictor of integration difficulty. Let $s_B$ denote the per-head attention score that position $A$ assigns to a preceding position $B$, scaled by the usual softmax temperature $\beta$ (the explicit form for per-head attention is given in Appendix~\ref{app:components}). Let $p$ denote the resulting attention distribution over preceding positions, i.e.\ the causal softmax with entries $p_B \propto \exp(s_B)$. As we noted, one energy component, $E^{\textit{AT}}_A(g)$, is the negated log-sum-exp over the attention scores.  Because $p$ is the softmax of $s$, the log-sum-exp can be decomposed (for a single head) into
\begin{equation}
\mathrm{LSE}(s) = \langle s\rangle_p + H(p),
\end{equation}
with the expected attention score $\langle s\rangle_p = \sum_B p_B s_B$ and the attention entropy $H(p) = -\sum_B p_B \log p_B$. Tying this to the attention energy, we thus see that $E^{\textit{AT}}_A$ decomposes into two pieces with distinct interpretations: the negative of the expected score, $-\langle s\rangle_p$, capturing how well the keys the model attends to match the query at $A$ on average, and a negated attention entropy, $-H(p)$ capturing how spread-out the attention is over the preceding context. The attention-entropy predictor of \citet{ryu2025} is, up to a scaling parameter, one of these two terms.

A subtlety must be flagged here. It is tempting to read the decomposition as ``energy contains, as one addend, attention entropy, so naturally, it will track its effects''. The situation is more complicated, because the two terms enter $E^{\textit{AT}}_A$ with opposite signs of contribution to difficulty. \citet{ryu2025} find that \emph{higher} attention entropy predicts \emph{more} difficulty (longer reading times), but in the decomposition higher attention entropy makes the attention energy, $E^{\textit{AT}}_A$ \emph{more negative}, i.e.\ lower. The expected-score term, by contrast, has the intuitive sign: when no preceding key matches the query well, the expected score is low and the energy is high. Because hard-to-integrate positions tend to have both low expected scores and spread-out attention, the two terms are correlated across items, but they push energy in opposite directions. Energy can therefore track difficulty effects via the expected-score term, with the attention-entropy term partially offsetting it; which term dominates an observed energy gap is an empirical question that we return to when interpreting our experiments.

\section{Experiments}

\subsection{Models}

We consider one NRGPT model with two training regimes. The model had 6 layer iterations, 12 heads, context length 1,024 tokens, embedding dimension 1,536. It was trained on OpenWebText  \citep{gokaslan2019openwebtext}. The parameters used in the model were found using grid search in \cite{nrgpt}, which showed that the trained model is competitive with GPT-2 and GPT-recursive-parallel models. For more details on the model parameters, see Appendix~\ref{app:params}.

We checked the model with two training regimes: one model trained in \cite{nrgpt}, downloaded from Hugging Face,\footnote{\url{https://huggingface.co/bsaha205/NRGPT-H-FF2W-128M-OWT}} which was trained for 500,000 iterations. Next to it, we also trained our own model with the same parameters but only for 100,000 iterations. The latter model is tested for practical reasons. Its training required around 60 hours on four H100s, which is within reach of NLP and psycholinguistic researchers. We wanted to see whether this amount of training is already sufficient for a good fit to psycholinguistic data. Such information is hopefully useful for researchers who want to develop and test their own energy-based models on human sentence processing. The results of the 100,000-iteration model are very similar to the model trained for 500,000 iterations and we do not discuss them further in the main text, but interested readers can find them in Appendix~\ref{app:smaller}.\footnote{Code for the experiments available at \url{https://github.com/jakdot/energy-transformers-reading-difficulty}.}

\subsection{Experiment 1: Relative Clause Processing}

As we noted, surprisal theory and attention entropy do not track the same effects in processing. If we claim that energy in NRGPT can potentially subsume both, we should look at instances where one, but not the other, is the dominant explanation.

One such well-known case concerns relative clauses. Consider the following pair of subject relative clause (SRC) and object relative clause (ORC):

\ex.
\a. The firemen that \textbf{called} the residents attacked the house with high-powered hoses. (SRC)
\b. The firemen that the residents \textbf{called} attacked the house with high-powered hoses. (ORC)

It has been shown that the embedded verb is processed slower in ORC, \Last[b], compared to  SRC, \Last[a] \citep{grodner2005, staub2010, levy13}. Surprisal theory has a hard time explaining this effect. Object relative clauses are less frequent than subject relative clauses, but higher surprisal should be present already at the noun phrase, \textit{the residents}, in \Last[b], since that already signals the ORC. At the verb, \Last[b] should be unambiguously ORC, so surprisal should play little role there. Indeed, it has been shown that the surprisal collected from GPT-2 does not show a difference between the verbs in SRC and ORC, contrary to human sentence processing. This is a point made clearly in \cite{ryu2025}, who furthermore show that attention entropy predicts processing difficulties for the embedded verb in ORC compared to SRC. On the other hand, \citet{staub2010} shows that the noun phrase \textit{the residents} is harder to process as the subject in ORC than the object in SRC, in line with surprisal, but not predicted by attention entropy \citep{ryu2025}. We explore how energy in NRGPT captures the contrasts in SRC and ORC.

\paragraph{Materials.}
We used the 24 sentence pairs from \citet{staub2010}, which contrast SRCs and ORCs matched for argument structure. To ensure equal token counts across conditions, we modified a subset of items by adding tokens at the start of the sentence, following the procedure of \citet{ryu2025}. Representative examples are given in \ref{ex:verb} and \ref{ex:the}.

\ex. Embedded verb contrasts:\label{ex:verb}
\a. Yesterday at early afternoon, the firemen that \textbf{called} the residents attacked the house with high-powered hoses. (SRC)
\b. Yesterday afternoon, the firemen that the residents \textbf{called} attacked the house with high-powered hoses. (ORC)

\ex. Noun phrase contrasts:\label{ex:the}
\a. Yesterday early afternoon, the firemen that called \textbf{the} residents attacked the house with high-powered hoses. (SRC)
\b. Yesterday very early afternoon, the firemen that \textbf{the} residents called attacked the house with high-powered hoses. (ORC)

\paragraph{Analysis.}

\begin{figure}[h]
  \includegraphics[width=\linewidth]{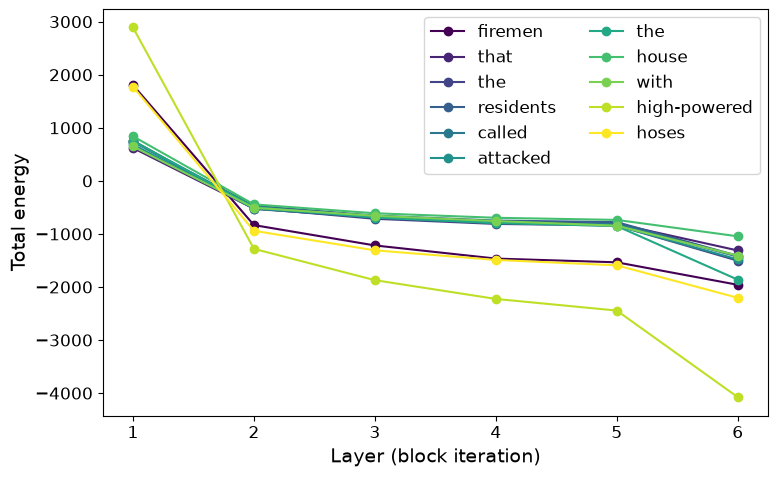}
  \caption{Energy trajectories per word across layer iterations in a relative clause. \label{fig1}}
\end{figure}

\begin{figure*}[h]
  \includegraphics[width=\textwidth]{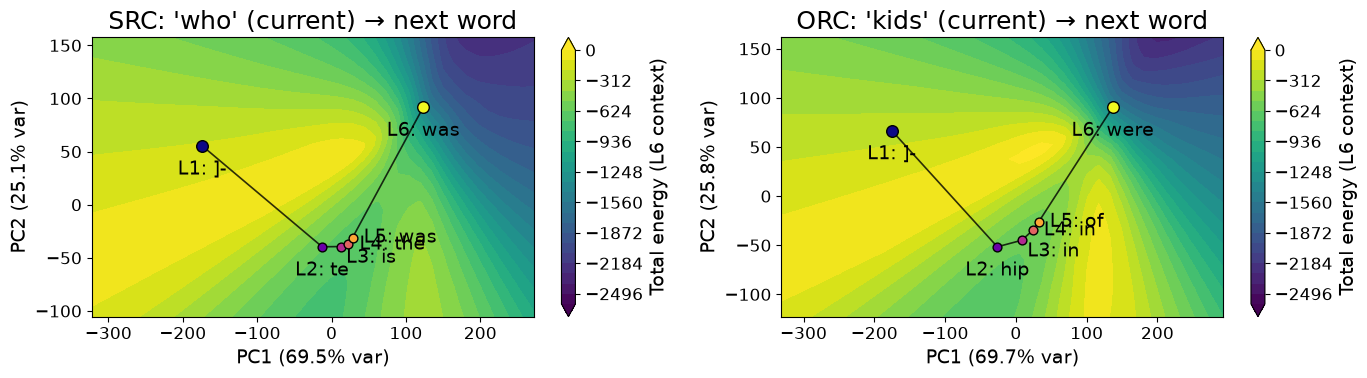}
  \includegraphics[width=\textwidth]{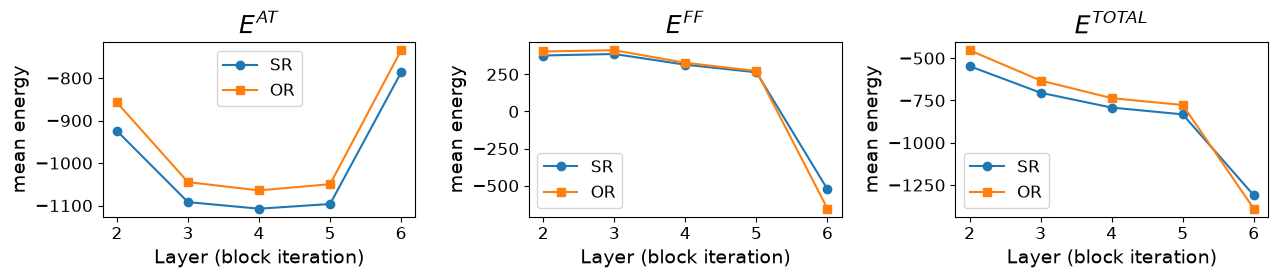}
  \includegraphics[width=\textwidth]{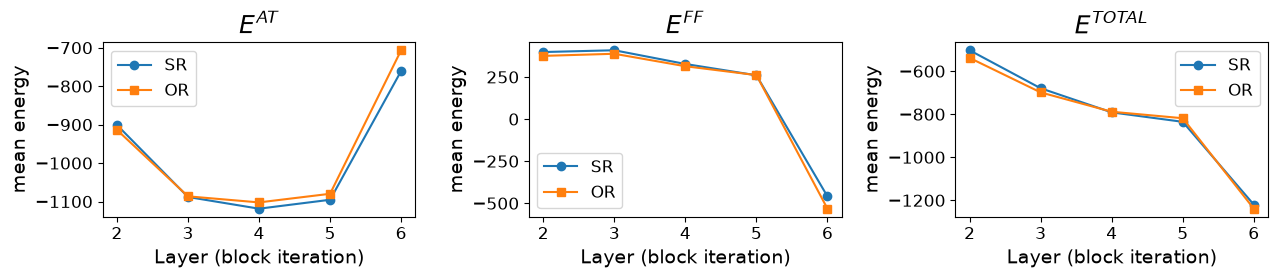}
  \caption{\textbf{Top:} Gradient descent in the context of the 6th layer for one example of the subject relative (SR, left) and the object relative (OR, right) clause. L1\ldots L6 - layer iterations right before the critical word (the verb). The predicted words next to the labels show the projection of LM head per layer. Middle and bottom: Energy trajectory across layer iterations for the embedded verb (top) and noun phrase onset (bottom) in SRC vs.\ ORC conditions. Each tick on the $x$-axis corresponds to one layer iteration in NRGPT.\label{fig2}}
\end{figure*}

While energies from various layers might be informative, we want to use the one that is close to converging, i.e., where the energy descent is stabilizing. We explore relative-clause examples and see that across words, the fifth layer iteration is the one where energy descent flattens out, followed by a bigger jump in the sixth iteration; see Fig.~\ref{fig1} and the top of Fig.~\ref{fig2}. We therefore focus on the last two layers: energy values in the fifth and the sixth layer iterations.

We evaluate two contrasts: the embedded verb, \ref{ex:verb}, and \textit{the} in the embedded noun phrase, \ref{ex:the}. For each position, we compare surprisal, energy at the two layers, and its decomposition into $E^{\textit{FF}}$, expected score and attention entropy between conditions across the 24 items. Just like for surprisal, which calculates, for the word at position $n$, the predictive measure for the word $n+1$, we think of energy and derived measures causally: we study whether energy for the word at position $n$ is predictive for the word $n+1$, i.e., the following word.

\paragraph{Results: embedded verb.}
For the embedded verb, surprisal of NRGPT does not differentiate the two conditions ($t = 0.1$), consistent with previous findings for GPT-2 \citep{ryu2025}. Energy at layer 5 predicts the contrast in the expected direction: the mean energy difference (ORC minus SRC) is 55 units, a highly significant effect ($t = 15.3$, $p \ll .001$). As shown in Figure~\ref{fig2}, the effect peaks at the fifth layer iteration and is nullified at the sixth. In fact, energy at layer 6 does not reach significance.

We break down the possible cause of the highly significant effect of energy in layer 5. Attention entropy in that layer also predicts the effect (ORC minus SRC: 3.9, $t=12.8, p\ll .001$), but recall that attention entropy has the opposite sign in the energy calculation. The effect in energy is instead driven by the (opposite sign of the) expected score in layer 5 (ORC minus SRC: -50.6, $t = -9.3, p\ll .001$) and $E^{\textit{FF}}$ (ORC minus SRC: 8.7, $t = 2.7, p< .01$).

\paragraph{Results: embedded noun phrase onset.}
At the embedded noun phrase onset (\textit{the}), surprisal predicts ORC difficulty strongly (ORC minus SRC: 4.5 nats, $t = 15.8$). Energy at layer 5 also predicts this contrast (ORC minus SRC: 16.7 units, $t = 4.7$, $p < .001$). Just as in the previous case, energy at layer 6 does not reach significance. Breaking down the effect of layer 5 energy, we see that attention entropy does not predict this effect (ORC minus SRC: -0.06, $t=-0.3$, $p>.1$). The effect is driven by the (opposite sign of the) expected score (ORC minus SRC: -15.5, $t = -2.8$, $p<.01$).

Energy is thus a significant predictor at both difficulty loci in relative-clause processing. Surprisal predicts only the embedded-subject effect; attention entropy predicts only the embedded-verb effect. Energy predicts both.

\subsection{Experiment 2: Reading-time corpora}

We study whether the energy measure in NRGPT can be used to predict reading times in psycholinguistic corpora. We compare the effect of energy (collected again at the fifth and sixth layers) to surprisal and attention entropy, collected from two models: NRGPT (for the closest comparison to energy) and GPT-2-small \citep{radford2019}. The latter is used for closer comparison with previous research. Various variants of attention-entropy measures have been proposed in literature (see \citealt{ryu2025, oh2022}). In the main text, we focus on raw attention entropy. Other variants, discussed in detail in \cite{oh2022}, are presented in Appendix~\ref{app:otherae}. Our conclusions and support for energy are independent of the chosen variant of attention-entropy measure.

\paragraph{Data.}
We use two corpora: the Natural Stories (NSC) corpus \citep{futrell2021}, which consists of 10 texts read in a self-paced reading paradigm; and eye-tracking and self-paced reading data from the UCL corpus \citep{UCLCorpus}. From both corpora, we excluded reading times shorter than 100 ms and greater than 3,000 ms. We also excluded reading times for an NSC story if the participant answered 4 or fewer out of 6 comprehension questions correctly. From the UCL, we excluded reading times for participants who answered less than 80 per cent of comprehension questions correctly. We calculated energy and surprisal measures on the corpora. The stories in the NSC range from 939 to 1,099 words, but the context window of NRGPT is 1,024 tokens, which corresponds to 850 words on average. We present the results for each story up to the context size, starting from the first word. In Appendix~\ref{app:sliding}, we show that a sliding window approach yields the same results. For multi-token words, we use the sum surprisal value and mean energy/entropy.\footnote{To ensure the effects are not driven by this decision, we also show results for single-token words in Appendix~\ref{app:single}. These largely show the same pattern as the results reported here.} We fit linear mixed-effects models predicting log-transformed reading times, following standard practice in the literature. The models were constructed on the 704k observation points for the NSC corpus, 37k for the UCL eye-tracking corpus, and 218k observations for the UCL self-paced reading corpus.

\paragraph{Models.}
We fit more than twenty linear mixed-effects models. All models are compared to the baseline model, which has fixed-effect covariates for word position in the text, log unigram frequency, log bigram frequency, and word length, and by-participant and by-story (by-sentence in the case of UCL) random intercepts. The models add to the covariates fixed-effect predictors of main interest, namely, surprisal (from NRGPT or GPT-2), entropy (from NRGPT or GPT-2 and from the fifth and sixth layer for NRGPT, and for all layers and the last, twelfth layer, for GPT-2), and energy (from the fifth or the sixth layer), and any combination of any two parameters. We are interested in the predictive effect of energy especially in addition to either surprisal and entropy. The dependent variable was reading time based on key-press duration (\textit{spr data}) and go-past reading time (\textit{eye tracking}), both log-transformed. In Appendix~\ref{app:reading_et}, we also report results for other reading-time eye-tracking measures, which are in line with the main findings reported here.
\begin{table*}[h]
  \centering
\begin{tabular}{lcrcrcr}
  \hline
  \textbf{Model with} & +/-/0? & $\mathbf{\Delta}$\textbf{LL} & +/-/0? & $\mathbf{\Delta}$\textbf{LL} & +/-/0? & $\mathbf{\Delta}$\textbf{LL}\\
  & NSC & NSC & UCL & UCL & UCL & UCL\\
  &  &  & (et) & (et) & (spr) & (spr)\\
  \hline
  Surprisal & + & $552$ & + & $155$ & 0 & $1$ \\
  Surprisal-GPT2 & + & $561$ & + & $153$ & - & $4$ \\
  Energy (L5) & + & $203$ & + & $8$ & + & $28$ \\
  Energy (L6) & - & $6$ & + & $386$ & - & $5$ \\
  Surprisal + Energy (L5) & + , + & $687$ & + , + & $166$ & 0 , + & $28$ \\
  Surprisal + Energy (L6) & + , - & $557$ & + , + & $487$ & 0 , - & $5$ \\
  Surprisal-GPT-2 + Energy (L5) & + , + & $\mathbf{695}$ & + , + & $166$ & - , + & $30$ \\
  Surprisal-GPT-2 + Energy (L6) & + , - & $568$ & + , + & $458$ & - , - & $7$ \\
  Entropy (L5) & + & $2$ & - & $82$ & + & $39$ \\
  Entropy (L6) & - & $28$ & - & $37$ & + & $2$ \\
  Surprisal + Entropy (L5) & + , 0 & $553$ & + , - & $246$ & 0 , + & $40$ \\
  Surprisal + Entropy (L6) & + , - & $594$ & + , - & $206$ & 0 , + & $3$ \\
  Surprisal-GPT-2 + Entropy (L5) & + , 0 & $562$ & + , - & $234$ & - , + & $43$ \\
  Surprisal-GPT-2 + Entropy (L6) & + , - & $606$ & + , - & $201$ & - , + & $6$ \\
  Entropy (L5) + Energy (L5) & - , + & $213$ & - , + & $136$ & + , + & $\mathbf{47}$ \\
  Entropy (L6) + Energy (L6) & - , - & $34$ & - , + & $410$ & + , - & $7$ \\
  Entropy-GPT2 (L12) & - & $38$ & - & $167$ & + & $20$ \\
  Surprisal + Entropy-GPT2 (L12) & + , - & $605$ & + , - & $306$ & 0 , + & $21$ \\
  Surprisal-GPT-2 + Entropy-GPT2 (L12) & + , - & $614$ & + , - & $292$ & - , + & $23$ \\
  Entropy-GPT2 (all layers) & - & $90$ & - & $421$ & + & $18$ \\
  Surprisal + Entropy-GPT2 (all layers) & + , - & $673$ & + , - & $\mathbf{540}$ & 0 , + & $18$ \\
  Surprisal-GPT-2 + Entropy-GPT2 (all layers) & + , - & $686$ & + , - & $514$ & 0 , + & $19$ \\
  Entropy-GPT2 (all layers) + Energy (L5) & - , + & $350$ & - , + & $464$ & + , + & $36$ \\
  Entropy-GPT2 (all layers) + Energy (L6) & - , - & $115$ & - , + & $539$ & + , 0 & $18$ \\
  \hline
\end{tabular}

  \caption{Mixed-effects model results for the Natural Stories (NSC) corpus and the self-paced reading (spr) and eye-tracking (et) portions of the UCL corpus. $\Delta$LL=difference between the log-likelihood of the models and the covariate-only baseline. The higher the $\Delta$LL value is, the better the fit of the model to the data. +/-/0: is the estimate of the psycholinguistic parameter(s) in the model positive (+), negative (-), or non-significant (0)? Linear mixed-effects models lack exact degrees of freedom to calculate precise $p$-values. As is common, we take values $|t|>1.96$ as significant. \label{tab:ns}}
\end{table*}

\paragraph{Results.}
Results are summarized in Table~\ref{tab:ns}. The table shows the difference of log likelihoods of the model of interest compared to the covariate-only baseline. The higher the number, the more variance is captured by the model. They also include the direction of the predictor effect, i.e., whether it was positive (+) or negative (-). 0 indicates non-significance. We expect surprisal, energy and entropy to be positively correlated with reading times. This is not always the case: surprisal of GPT-2 is negative in the UCL self-paced reading data and layer 6 energy appears as a negative predictor in NSC and UCL (spr) data. Entropy has the wrong (negative) sign in UCL (et) data and in some cases of the NSC data. Layer 5 energy always goes in the expected direction. 

Surprisal provides a better fit than energies in the NSC corpus, while energies provide a better fit in the UCL (et) and UCL (spr) corpora. Energies furthermore remain significant predictors when combined with surprisal into one model. Similar conclusions can be drawn for the comparison between energy and entropy. Energy L5 shows a greater increase in $\Delta$LL than attention entropy in NSC. Furthermore, energies remain significant predictors when combined with attention entropy. Energy thus accounts for substantial variance in reading times over and above surprisal and over and above attention entropy.

The winning models for NSC self-paced reading and UCL self-paced reading are those that include energy -- which either combines with entropy, or surprisal, to deliver the best fit. In case of the UCL (et) corpus, the best model is Surprisal + Entropy-GPT2 (all layers). However, importantly, entropy has the wrong sign in this case, and thus, this finding, if anything, is problematic for entropy-based accounts of reading. The best model for UCL eye-tracking in which the predictors have the expected sign, is again the one with energy: Surprisal + Energy (L6). In sum, energy seems to be the crucial ingredient for the best models across all the considered datasets.

As we noted before, the energy consists of two components, one of which, attention energy, consists of two parts in turn: expected score and attention entropy. In Appendix~\ref{app:breakdown}, we provide a detailed breakdown of energy components to indicate which of them play a crucial role in the model fit to the data.

\section{Discussion}

Across the experiments, we see that the energy of NRGPT is a robust predictor of reading difficulty. Energy at the fifth layer is particularly predictive of the reading-time patterns. In our experiments on corpora, we see that energy is a strong predictor of self-paced reading data. The models with the fifth-layer energy are the winning ones, combining either with surprisal or with entropy as the second variable of interest. In our Experiment~1 on relative clauses, we see that energy can capture two effects for which both surprisal and attention entropy are normally necessary. The latter finding provides empirical evidence that in addition to predictive processing captured by surprisal, the fifth-layer energy also captures memory-related aspects of processing. It remains to be seen whether other attempts to limit transformer models with human-like memory constraints to improve the fit of surprisal to reading-time data, for instance, by using lossy-context surprisal \citep{futrell2017noisy, futrell2020, hahn2022}, align with our research.

Interestingly, the model with energy of the sixth layer provides the best fit for the UCL eye-tracking data. It might be tempting to assume that the fifth layer energy is more representative for self-paced reading data and the sixth layer energy is more representative for eye-tracking data, but we have to note that the UCL eye-tracking data have far fewer observations than the self-paced reading data, so one should be careful about the interpretation of these differences.

In our studies, we mainly focused on the whole energy of a layer. As we noted, the total energy that we considered has two components: the feedforward energy, $E^{\textit{FF}}$, and the attention energy, $E^{\textit{AT}}$. The latter, in turn, can be decomposed into (scaled) expected attention score and attention entropy. We saw that expected attention score was particularly useful in fitting reading-time data. This result is relevant also for researchers who work with classical transformers and are not interested in energy models. While researchers have proposed that there is an intuitive link between cue-based retrieval memory and attention entropy \citep{ryu2025}, expected attention score, to the best of our knowledge, has not been studied in sentence processing from this perspective. Based on our results, it seems possible that the latter measure would be a better proxy for reading-time data. It remains to be seen whether this also holds for classical transformers.

\section{Conclusion}

We investigated the energy function of NRGPT as a predictor of human reading difficulty, to the best of our knowledge the first such use of an energy-based transformer in computational psycholinguistics. The energy scalar provides a principled link to associative memory models and, as our results show, a practically useful predictor of reading times. Across three corpora, NRGPT energy is a robust predictor that contributes significantly over and above surprisal and attention entropy. Furthermore, a focused case study on the processing of relative clauses shows that NRGPT energy captures aspects of reading for which normally both surprisal and attention entropy are necessary. We hope this work encourages further exploration of energy-based transformers in psycholinguistics and motivates closer theoretical ties between sentence processing research and computational research on associative memory.

\section*{Limitations}

This study is a first exploration of energy-based transformers as predictors of human sentence processing, and its scope is correspondingly narrow in several respects. First, we study a single model in just one configuration and at two training checkpoints. We do not know how our findings generalize to other energy-based transformers, to other architectural choices within NRGPT, or to other model sizes. Given that classical transformers show systematic shifts in reading-time fit as model size and training change, similar dependencies are plausible here and remain to be mapped out.

Second, our empirical base is limited. We evaluate on three reading-time corpora (Natural Stories, UCL eye-tracking, UCL self-paced reading) and one controlled relative-clause manipulation. Broader testing across languages, constructions, and reading paradigms would be needed before drawing strong conclusions about the generality of energy as a processing predictor.

Third, we examine only selected measures derived from the model: the total layer energy and its decomposition into expected attention score and attention entropy, compared against surprisal. Other measures could be derived from the energy landscape (for example, the number of descent steps to convergence, or properties of the trajectory), and these may carry additional processing-relevant information that we have not explored.

Finally, we restrict the dependent measure to reading times. Other behavioral and neural measures, such as brain activation from fMRI or EEG, are an important target for future work and would provide a stronger test of whether energy tracks processing difficulty in general rather than reading time specifically.

\section*{Ethical statement}

This work relies entirely on previously published, publicly available resources. The behavioral data come from the Natural Stories corpus \citep{futrell2021} and the UCL corpus \citep{UCLCorpus}, both of which were collected and released by the original authors under appropriate ethical approvals and participant consent. We did not collect any new data from human participants. The data we use contain no personally identifying information.

The models we use, NRGPT \citep{nrgpt} and GPT-2 \citep{radford2019}, and their training data are publicly available and were used in accordance with their licenses and intended research use. We foresee no harmful applications or risks to individuals arising from this work.

\section*{Acknowledgments}

The research reported in this paper was supported by the European Research Council (ERC), grant 101088098 - MEMLANG. Views and opinions expressed are however those of the authors only and do not necessarily reflect those of the European Union or the European Research Council Executive Agency. Neither the European Union nor the granting authority can be held responsible for them. We thank the members of the MEMLANG group for their comments.

LLM assistants were used in circumscribed and local, and closely supervised ways during code development and the editing process of the manuscript. The authors retain full responsibility for all content, and all errors remain our own.

\bibliography{custom}

\appendix

\section{Energy, attention and feedforward components}
\label{app:components}

This appendix shows in more detail how the two energy components of NRGPT are designed so that their gradients reproduce, up to a reparametrization of the weights, the standard attention and feedforward operations of transformer models. Throughout, $g$ denotes the normalized token states, i.e.\ $g = \mathrm{LN}(x)$ where $\mathrm{LN}$ is a normalization operation (LayerNorm or RMSNorm); we refer the reader to \citet{nrgpt} for the precise form and its role in the stability of the dynamics. For multi-head attention with $H$ heads and hidden per-head dimension $Y=D/H$, with query and key projections $W^Q, W^K\in \mathbb{R}^{H\times Y\times D}$, the per-token attention energy at position $A$ takes the form
\begin{equation}
\begin{aligned}
  E^{\textit{AT}}_A&(g) \;=\;\\ &-\frac{1}{\beta}\sum_{h=1}^{H} \alpha_h\,\mathrm{LSE}_{B<A}\!\left(\beta \, g_B^\top (W_h^K)^\top W_h^Q \, g_A\right),
\end{aligned}
\end{equation}
with $\beta = 1/\sqrt{Y}$ and a learnable per-head weight $\alpha_h$. Here $\mathrm{LSE}(z) = \log\sum_i \exp(z_i)$ is the log-sum-exp, and $\mathrm{LSE}_{B<A}$ restricts the sum to preceding positions $B < A$. This restriction encodes the causal mask, so $E^{\textit{AT}}_A$ depends on $g_A$ (the query) and on $g_B$ for $B < A$ (the keys). Taking the gradient of $E^{\textit{AT}}_A$ with respect to $g_A$ yields an expression familiar as the standard attention operation in the transformer block, namely,
\begin{equation}
\begin{aligned}
  \text{AT}&(g_A) = -\eta\,\frac{\partial E^{\textit{AT}}_A(g)}{\partial g_A} = \\
& \sum_{h=1}^{H} \alpha_h\,\eta\, (W_h^Q)^{\!\top} W_h^K\, g\;\text{SM}\!\left(g^{\!\top} (W_h^K)^{\!\top} W_h^Q\, g_A\right),
\end{aligned}
\label{eq:nrgpt-attn}
\end{equation}
where $\text{SM}\!\left(g^{\!\top} (W_h^K)^{\!\top} W_h^Q\, g_A\right)$ is the softmax over positions $B < A$, treated as a vector that multiplies the projected token states $\eta\,(W_h^Q)^{\!\top} W_h^K\, g$ in \eqref{eq:nrgpt-attn}. Its $B$-th entry is
\begin{equation}
\begin{aligned}
  \text{SM}&\!\left(g^{\!\top} (W_h^K)^{\!\top} W_h^Q\, g_A\right)_B = \\
  &\frac{\exp\!\left(\beta\, g_B^{\!\top} (W_h^K)^{\!\top} W_h^Q\, g_A\right)}{\sum_{C<A} \exp\!\left(\beta\, g_C^{\!\top} (W_h^K)^{\!\top} W_h^Q\, g_A\right)}.
\end{aligned}
\label{eq:nrgpt-sm}
\end{equation}

This has the same structure as a causal multi-head attention layer, with the conventional value-and-output projection $[W^P]^{\top} W^V$ replaced by $\alpha_h\,\eta\,(W_h^Q)^{\top} W_h^K$.

The feedforward energy is constructed analogously. The feedforward energy is a sum of per-token contributions in the dense associative memory form,
\begin{equation}
  E^{\textit{FF}}(g) = -\sum_{A=1}^{N} \mathbf{1}^\top F\!\left(W^1 g_A\right), \quad \text{s.t.}\ F' = \sigma,
\label{eq:nrgpt-eff}
\end{equation}
where $W^1 \in \mathbb{R}^{M \times D}$ is the first feedforward weight matrix mapping the per-token state $g_A$ into a hidden dimension $M$, $\mathbf{1}$ is an $M$-dimensional vector of ones, and $F$ is a scalar nonlinearity whose derivative is the activation $\sigma$. Because each $E^{\textit{FF}}_A$ depends only on $g_A$, differentiating the total energy with respect to $g_A$ picks out the $A$-th term and yields a two-layer feedforward sublayer,
\begin{equation}
\text{FF}(g_A) = -\eta\,\frac{\partial E^{\textit{FF}}}{\partial g_A} = \eta\, W^{1\top}\sigma\!\left(W^1 g_A\right).
\label{eq:nrgpt-ff}
\end{equation}
Both the standard MLP in transformers and the update in \eqref{eq:nrgpt-ff} have the form $(\,\cdot\,)\,\sigma(W^1 g_A)$. A standard MLP applies an independently learned second weight matrix $W^2 \in \mathbb{R}^{D \times M}$, whereas in \eqref{eq:nrgpt-ff} the effective second matrix is $\eta\, W^{1\top}$. $\eta$ is itself learnable. More general feedforward energies that retain a separate $W^2$ are also possible, and are in fact what we use in our experiments; see \citet{nrgpt} for further details.

In short, taking gradients of $E^{\textit{AT}}_A$ and $E^{\textit{FF}}_A$ gives back an attention sublayer and a feedforward sublayer, and one NRGPT iteration is therefore very close to one parallel-style transformer layer with weights tied across iterations.

\section{Model parameters and packages}
\label{app:params}

We use the \texttt{NRGPT\_H\_FF2W} variant of \citet{nrgpt} (energy attention combined with the two-weight feedforward energy), trained on OpenWebText. The architecture and training configuration are summarized in Table~\ref{tab:params}; all values are taken from \citet{nrgpt} (their Table 5 for the model-specific settings and Table 7 for the shared OpenWebText hyperparameters). The two models in our experiments share this configuration and differ only in the number of training iterations: 100{,}000 for the model we trained ourselves and 500{,}000 for the publicly released checkpoint.

\begin{table}[h]
  \centering
  \begin{tabular}{ll}
    \hline
    \textbf{Parameter} & \textbf{Value} \\
    \hline
    \multicolumn{2}{l}{\textit{Architecture}} \\
    Variant & \texttt{NRGPT\_H\_FF2W} \\
    Embedding dim. & 1{,}536 \\
    Layer iterations & 6 \\
    Attention heads & 12 \\
    Context length & 1{,}024 \\
    Feedforward hidden dim. & $4\times$ embed.\ (6{,}144) \\
    Normalization & LayerNorm \\
    Tokenizer & GPT-2 BPE (50{,}257) \\
    Total parameters & 128M \\
    \hline
    \multicolumn{2}{l}{\textit{Training (OpenWebText)}} \\
    Optimizer & AdamW \\
    $(\beta_1, \beta_2)$ & $(0.9,\ 0.99)$ \\
    Learning rate & $3\!\times\!10^{-5}$ \\
    LR schedule & cos decay to $3\!\times\!10^{-6}$ \\
    Weight decay & $10^{-2}$ \\
    Batch size & 12 \\
    Grad.\ accumulation steps & 40 \\
    Warmup iterations & 2{,}000 \\
    Dropout & 0.0 \\
    Training iterations & 100{,}000 / 500{,}000 \\
    \hline
  \end{tabular}
  \caption{Architecture and training configuration of the NRGPT model used in our experiments. \label{tab:params}}
\end{table}

For all experiments, we used Python 3.12 with PyTorch version 2.8.0 and transformers 4.46.3. For the mixed-effects model analysis, we used R 4.3.3 with lme4 version 1.1.

\section{Other eye-tracking reading measures}
\label{app:reading_et}

\begin{table*}[h]
  \centering
\begin{tabular}{lcrcrcr}
  \hline
  & \multicolumn{2}{c}{First fixation} & \multicolumn{2}{c}{First pass} & \multicolumn{2}{c}{Right-bounded} \\
  \textbf{Model with} & +/-/0? & \textbf{$\Delta$LL} & +/-/0? & \textbf{$\Delta$LL} & +/-/0? & \textbf{$\Delta$LL} \\
  \hline
  Surprisal & + & $38$ & + & $132$ & + & $177$ \\
  Surprisal-GPT2 & + & $50$ & + & $163$ & + & $194$ \\
  Energy (L5) & - & $14$ & - & $42$ & - & $23$ \\
  Energy (L6) & + & $183$ & + & $356$ & + & $299$ \\
  Surprisal + Energy (L5) & + , - & $51$ & + , - & $169$ & + , - & $196$ \\
  Surprisal + Energy (L6) & + , + & $\mathbf{204}$ & + , + & $\mathbf{441}$ & + , + & $\mathbf{425}$ \\
  Surprisal + Entropy (L5) & + , - & $70$ & + , - & $199$ & + , - & $242$ \\
  Surprisal + Entropy (L6) & + , - & $56$ & + , - & $171$ & + , - & $218$ \\
  Surprisal-GPT-2 + Energy (L5) & + , - & $62$ & + , - & $195$ & + , - & $209$ \\
  Surprisal-GPT-2 + Energy (L6) & + , + & $203$ & + , + & $438$ & + , + & $411$ \\
  Entropy (L5) & - & $29$ & - & $59$ & - & $57$ \\
  Entropy (L6) & - & $13$ & - & $28$ & - & $28$ \\
  Surprisal-GPT-2 + Entropy (L5) & + , - & $79$ & + , - & $222$ & + , - & $251$ \\
  Surprisal-GPT-2 + Entropy (L6) & + , - & $67$ & + , - & $201$ & + , - & $233$ \\
  Entropy (L5) + Energy (L5) & - , - & $32$ & - , - & $72$ & - , - & $61$ \\
  Entropy (L6) + Energy (L6) & - , + & $191$ & - , + & $373$ & - , + & $316$ \\
  Entropy-GPT2 (L12) & - & $53$ & - & $148$ & - & $116$ \\
  Surprisal + Entropy-GPT2 (L12) & + , - & $88$ & + , - & $268$ & + , - & $279$ \\
  Surprisal-GPT-2 + Entropy-GPT2 (L12) & + , - & $95$ & + , - & $285$ & + , - & $284$ \\
  Entropy-GPT2 (L12) + Energy (L6) & 0 , + & $184$ & - , + & $367$ & - , + & $305$ \\
  Entropy-GPT2 (all layers) & - & $84$ & - & $223$ & - & $221$ \\
  Surprisal + Entropy-GPT2 (all layers) & + , - & $115$ & + , - & $331$ & + , - & $369$ \\
  Surprisal-GPT-2 + Entropy-GPT2 (all layers) & + , - & $119$ & + , - & $341$ & + , - & $365$ \\
  Entropy-GPT2 (all layers) + Energy (L6) & - , + & $191$ & - , + & $397$ & - , + & $350$ \\
  \hline
\end{tabular}

\caption{Results using NRGPT and GPT-2 energy/entropy/surprisal values on other reading time measures from UCL eye-tracking corpus: first fixation, first pass, right-bounded reading times. See the text for the explanation of these reading measures.  Format as in Table~\ref{tab:ns}.} \label{tab:otherrt}
\end{table*}

Table~\ref{tab:otherrt} shows results for other reading-time measures from the UCL eye-tracking corpus. These are first-fixation time (duration of only the first fixation on the read word), first-pass time (also known as gaze duration: summed duration of all fixations on the read word before the
first fixation on any other word), right-bounded time (summed duration of all fixations on the read word before the first fixation on a word further to the right). These results replicate the main findings on the go-past reading, reported in the main text: the model with Energy (L6) has a better fit than the Surprisal model and than the Entropy model. The best model is the one combining Surprisal and Energy (L6).

\section{Breakdown of results per energy type\label{app:breakdown}}

We present the breakdown of corpus results per energy type for the winning models.

In case of the NSC data, the winning model includes Surprisal and Energy (L5). The positive energy effect is driven by both energy components: the positive effect of $E^{\textit{FF}}$ ($t=6.3$) and the positive effect of $E^{\textit{AT}}$ ($t=13.5$). Breaking down $E^{\textit{AT}}$ further, we see that this is due to the negative effect of expected attention score ($t=-14.3$) and the negative effect of attention entropy ($t=-9.2$).

The winning model for the UCL (et) data includes Surprisal and Energy (l6). The positive energy effect is driven by both energy components: the positive effect of $E^{\textit{FF}}$ ($t=16.8$) and the positive effect of $E^{\textit{AT}}$ ($t=14.3$). Breaking down $E^{\textit{AT}}$ further, we see that this is due to the negative effect of expected attention score ($t=-15.1$) and the negative effect of attention entropy ($t=-8.5$).

In case of the UCL (spr) data, the winning model includes Entropy (L5) and Energy (L5). The positive effect of Energy (L5) is driven by the positive effect of $E^{\textit{FF}}$ ($t=8.0$) and the positive effect of $E^{\textit{AT}}$ ($t=2.9$). Breaking down $E^{\textit{AT}}$ further, we see that its effect is due to the negative effect of expected attention score ($t=-2.9$), which, however, is offset by the positive effect of attention entropy ($t=7.1$).

\section{NSC results with sliding window\label{app:sliding}}

The NSC data in the main text were cut off to fit the context of the NRGPT model (1,024 tokens). For completeness, we include modeling results using the sliding window. We fit the second window so that the end of each story would correspond to the end of the context window. Each story was treated as separate. The results for NSC are summarized in Table \ref{tab:sliding}. The best model is the same as the best model without the sliding window: the model with Surprisal-GPT-2 and Energy (L5).

\begin{table*}[h]
  \centering
\begin{tabular}{lcr}
  \hline
  \textbf{Model with} & +/-/0? & $\mathbf{\Delta}$\textbf{LL}\\
  & NSC & NSC\\
  \hline
  Surprisal & + & $642$ \\
  Surprisal-GPT2 & + & $698$ \\
  Energy (L5) & + & $459$ \\
  Energy (L6) & 0 & $1$ \\
  Surprisal + Energy (L5) & + , + & $1{,}019$ \\
  Surprisal + Energy (L6) & + , + & $644$ \\
  Surprisal-GPT-2 + Energy (L5) & + , + & $\mathbf{1{,}067}$ \\
  Surprisal-GPT-2 + Energy (L6) & + , 0 & $699$ \\
  Entropy (L5) & - & $8$ \\
  Entropy (L6) & - & $49$ \\
  Surprisal + Entropy (L5) & + , - & $672$ \\
  Surprisal + Entropy (L6) & + , - & $709$ \\
  Surprisal-GPT-2 + Entropy (L5) & + , - & $727$ \\
  Surprisal-GPT-2 + Entropy (L6) & + , - & $770$ \\
  Entropy (L5) + Energy (L5) & - , + & $469$ \\
  Entropy (L6) + Energy (L6) & - , 0 & $50$ \\
  Entropy-GPT2 (L12) & - & $106$ \\
  Surprisal + Entropy-GPT2 (L12) & + , - & $780$ \\
  Surprisal-GPT-2 + Entropy-GPT2 (L12) & + , - & $834$ \\
  Entropy-GPT2 (all layers) & - & $270$ \\
  Surprisal + Entropy-GPT2 (all layers) & + , - & $977$ \\
  Surprisal-GPT-2 + Entropy-GPT2 (all layers) & + , - & $1{,}038$ \\
  Entropy-GPT2 (all layers) + Energy (L5) & - , + & $658$ \\
  Entropy-GPT2 (all layers) + Energy (L6) & - , - & $284$ \\
  \hline
\end{tabular}

\caption{Results using NRGPT and GPT-2 energy/entropy/surprisal values on NSC with the sliding window. See the text for the explanation of how the sliding window was applied. Format as in Table~\ref{tab:ns}.} \label{tab:sliding}
\end{table*}

\section{Variants on attention entropy\label{app:otherae}}

There are several variants on how to calculate attention entropy, discussed in \cite{ryu2025} and \cite{oh2022}. For completeness, we consider them in our reading-time corpus data, see Table~\ref{tab:otherae}. The abbreviations in the table are spelled out in the caption. See \cite{oh2022} for further details.

The first three rows in the table are the best models per corpus, as discussed in the main text. Importantly for us, these alternative attention-entropy measures do not change our main results. Two of the best models stay the best. Just as we observed before, the best model for UCL et is misleading, since the attention-entropy predictor carries the wrong sign. If we restrict our attention to only models in which the predictors of interest have the correct, expected sign, the original model (Surprisal + Energy (L6)) stays as the best here as well.

\begin{table*}[h]
  \centering
\begin{tabular}{lcrcrcr}
  \hline
  \textbf{Model with} & +/-/0? & $\mathbf{\Delta}$\textbf{LL} & +/-/0? & $\mathbf{\Delta}$\textbf{LL} & +/-/0? & $\mathbf{\Delta}$\textbf{LL}\\
  & NSC & NSC & UCL & UCL & UCL & UCL\\
  &  &  & (et) & (et) & (spr) & (spr)\\
  \hline
  Surprisal-GPT-2 + Energy (L5) & + , + & $\mathbf{695}$ & + , + & $166$ & - , + & $30$ \\
  Surprisal + Energy (L6) & + , - & $557$ & + , + & $487$ & 0 , - & $5$ \\
  Entropy (L5) + Energy (L5) & - , + & $213$ & - , + & $136$ & + , + & $\mathbf{47}$ \\\hline
  NAE (L5) & + & $36$ & - & $204$ & + & $27$ \\
  NAE (L6) & - & $14$ & - & $86$ & 0 & $1$ \\
  Surprisal + NAE (L5) & + , + & $562$ & + , - & $349$ & 0 , + & $28$ \\
  Surprisal + NAE (L6) & + , - & $577$ & + , - & $246$ & 0 , 0 & $2$ \\
  Surprisal-GPT-2 + NAE (L5) & + , + & $572$ & + , - & $325$ & - , + & $30$ \\
  Surprisal-GPT-2 + NAE (L6) & + , - & $588$ & + , - & $238$ & - , 0 & $4$ \\
  NAE (L5) + Energy (L5) & 0 , + & $204$ & - , + & $337$ & + , + & $37$ \\
  NAE (L6) + Energy (L6) & - , - & $19$ & - , + & $417$ & 0 , - & $5$ \\
  NAE (L12) & - & $6$ & - & $349$ & + & $15$ \\
  Surprisal + NAE (L12) & + , - & $568$ & + , - & $447$ & 0 , + & $15$ \\
  Surprisal-GPT-2 + NAE (L12) & + , - & $577$ & + , - & $423$ & 0 , + & $16$ \\
  NAE (L12) + Energy (L6) & - , - & $15$ & - , + & $429$ & + , 0 & $15$ \\
  NAE (all layers) & - & $3$ & - & $571$ & + & $11$ \\
  Surprisal + NAE (all layers) & + , - & $572$ & + , - & $650$ & 0 , + & $12$ \\
  Surprisal-GPT-2 + NAE (all layers) & + , - & $583$ & + , - & $620$ & 0 , + & $13$ \\
  NAE (all layers) + Energy (L5) & - , + & $250$ & - , + & $\mathbf{677}$ & + , + & $31$ \\
  NAE (all layers) + Energy (L6) & - , - & $12$ & - , + & $588$ & + , 0 & $11$ \\
  NAE-N (L12) & - & $12$ & - & $280$ & + & $16$ \\
  Surprisal-GPT2 + NAE-N (L12) & + , - & $581$ & + , - & $390$ & - , + & $18$ \\
  Energy (L5) + NAE-N (L12) & + , - & $229$ & + , - & $296$ & + , + & $36$ \\
  Energy (L6) + NAE-N (L12) & - , - & $20$ & + , - & $480$ & 0 , + & $16$ \\
  dNAE (L12) & + & $35$ & - & $78$ & + & $4$\\
  Surprisal-GPT2 + dNAE (L12) & + , + & $612$ & + , - & $209$ & - , + & $7$ \\
  Energy (L5) + dNAE (L12) & + , + & $230$ & + , - & $112$ & + , + & $32$ \\
  Energy (L6) + dNAE (L12) & - , + & $41$ & + , - & $397$ & - , + & $7$\\
  MD (L12) & 0 & $2$ & - & $82$ & + & $17$\\
  Surprisal-GPT2 + MD (L12) & + , 0 & $561$ & + , - & $196$ & 0 , + & $19$\\
  Energy (L5) + MD (L12) & + , - & $215$ & + , - & $124$ & + , + & $35$ \\
  Energy (L6) + MD (L12) & - , - & $9$ & + , + & $395$ & 0 , + & $18$\\
  dNAE-N (L12) & + & $66$ & - & $78$ & + & $4$\\
  Surprisal-GPT2 + dNAE-N (L12) & + , + & $649$ & + , - & $209$ & - , + & $7$ \\
  Energy (L5) + dNAE-N (L12) & + , + & $253$ & + , - & $88$ & + , + & $31$ \\
  Energy (L6) + dNAE-N (L12) & - , + & $72$ & + , - & $397$ & - , + & $7$ \\
  MD-N (L12) & + & $4$ & - & $108$ & + & $14$ \\
  Surprisal-GPT2 + MD-N (L12) & + , + & $573$ & + , - & $210$ & 0 , + & $15$ \\
  Energy (L5) + MD-N (L12) & + , 0 & $203$ & + , - & $153$ & + , + & $33$ \\
  Energy (L6) + MD-N (L12) & - , + & $9$ & + , + & $391$ & 0 , + & $14$ \\
  \hline
\end{tabular}

  \caption{Results of the corpus modeling with variants of attention-entropy measures. NAE=attention entropy normalized to the highest entropy attainable per position; NAE-N=NAE weighted by the L2-norm of the output vector; dNAE=difference in normalized attention entropy between consecutive words; MD=Manhattan (L1-norm) distance of normalized attention entropy between consecutive words; MD-N= Manhattan (L1-norm) distance using NAE-N, rather than NAE. L12 is the twelfth, i.e., the last layer of GPT-2, L5 and L6 are the fifth and sixth layer of NRGPT. The first three rows are the best model presented in the main text. The rest of the formatting follows Table~\ref{tab:ns}. \label{tab:otherae}}

\end{table*}

\section{Results: the 100,000 model}
\label{app:smaller}

\subsection{Experiment 1: Relative Clause Processing}

We repeat the relative-clause analysis of Experiment~1 with the model trained for only 100{,}000 iterations. The pattern of the main (500{,}000-iteration) model replicates: energy at layer~5 significantly predicts both difficulty contrasts in the expected direction (ORC $>$ SRC). At the embedded verb, the mean energy difference (ORC minus SRC) is 290 units ($t = 13.2$); at the embedded noun-phrase onset (\textit{the}), it is 69.2 units ($t = 3.7$). Both effects are highly significant. This indicates that 100{,}000 training iterations already suffice for NRGPT energy to capture the relative-clause processing contrasts.

\subsection{Experiment 2: Reading-time corpora}

\begin{table*}[h]
  \centering
  \begin{tabular}{lcrcrcr}
  \hline
  \textbf{Model with} & +/-/0? & $\mathbf{\Delta}$\textbf{LL} & +/-/0? & $\mathbf{\Delta}$\textbf{LL} & +/-/0? & $\mathbf{\Delta}$\textbf{LL}\\
  & NSC & NSC & UCL & UCL & UCL & UCL\\
  &  &  & (et) & (et) & (spr) & (spr)\\
  \hline
  Surprisal & + & $561$ & + & $321$ & 0 & $1$ \\
  Surprisal-GPT2 & + & $561$ & + & $153$ & - & $4$ \\
  Energy (L5) & + & $163$ & + & $39$ & 0 & $0$ \\
  Energy (L6) & 0 & $0$ & + & $48$ & - & $8$ \\
  Surprisal + Energy (L5) & + , + & $\mathbf{674}$ & + , + & $342$ & 0 , 0 & $1$ \\
  Surprisal + Energy (L6) & + , 0 & $562$ & + , + & $348$ & 0 , - & $10$ \\
  Surprisal-GPT-2 + Energy (L5) & + , + & $666$ & + , + & $174$ & - , 0 & $4$ \\
  Surprisal-GPT-2 + Energy (L6) & + , - & $563$ & + , + & $183$ & - , - & $\mathbf{11}$ \\
  Entropy (L5) & + & $4$ & - & $44$ & + & $6$ \\
  Entropy (L6) & - & $7$ & - & $28$ & 0 & $2$ \\
  Surprisal + Entropy (L5) & + , 0 & $561$ & + , - & $\mathbf{360}$ & 0 , + & $6$ \\
  Surprisal + Entropy (L6) & + , - & $571$ & + , - & $352$ & 0 , 0 & $3$ \\
  Surprisal-GPT-2 + Entropy (L5) & + , 0 & $562$ & + , - & $198$ & - , + & $9$ \\
  Surprisal-GPT-2 + Entropy (L6) & + , - & $571$ & + , - & $183$ & - , 0 & $5$ \\
  \hline
\end{tabular}

  \caption{Mixed-effects model results for the 100{,}000-iteration model. Format as in Table~\ref{tab:ns}. \label{tab:ns100k}}
\end{table*}

Table~\ref{tab:ns100k} replicates Experiment~2 with the 100{,}000-iteration model. The overall pattern largely mirrors the results of the 500{,}000-iteration model. In the NSC corpus, Surprisal + Energy (L5) is the best-fitting model, and energy is a strong predictor over and above surprisal.  For UCL self-paced reading, Energy (L6) provides the best single-predictor fit, and Surprisal-GPT-2 + Energy (L6) is the best-fitting combined model. However, note that the energy effect goes in the opposite direction from what is theoretically predicted. One difference from the main model is that Energy (L5) does not significantly improve fit over surprisal alone in the UCL spr corpus at this training stage. For UCL eye tracking, Surprisal + Entropy (L5) achieves the best fit, but note that the effect of entropy is negative, i.e., the increase in entropy correlates with the decrease in reading times, which goes in the opposite direction to what the theory predicts, and should not be considered as a successful case of the entropy account. If we limit ourselves to cases in which the psycholinguistic predictors go in the theoretically predicted direction, Surprisal + Energy (L6) is the best model.

\section{Results: single token}
\label{app:single}

See Table~\ref{tab:ns_single} for the results on the corpora limited to single-token words. These results largely align with the results that included multi-token words (Table~\ref{tab:ns}). One exception is UCL (spr). This dataset shows the positive effect of Energy (L5), but in this particular case, it also shows that the model with Surprisal-GPT-2 and Entropy (L5) is better fitting to the data. This is one instance in which the model with surprisal and entropy better fits the data than a model with energy and one of surprisal/entropy and entropy goes in the expected direction.

\begin{table*}[h]
  \centering
\begin{tabular}{lcrcrcr}
  \hline
  \textbf{Model with} & +/-/0? & $\mathbf{\Delta}$\textbf{LL} & +/-/0? & $\mathbf{\Delta}$\textbf{LL} & +/-/0? & $\mathbf{\Delta}$\textbf{LL}\\
  & NSC & NSC & UCL & UCL & UCL & UCL\\
  &  &  & (et) & (et) & (spr) & (spr)\\
  \hline
  Surprisal & + & $305$ & + & $60$ & 0 & $1$ \\
  Surprisal-GPT2 & + & $314$ & + & $73$ & 0 & $2$ \\
  Energy (L5) & + & $208$ & - & $53$ & + & $6$ \\
  Energy (L6) & - & $20$ & + & $103$ & - & $15$ \\
  Surprisal + Energy (L5) & + , + & $458$ & + , - & $114$ & 0 , + & $7$ \\
  Surprisal + Energy (L6) & + , - & $326$ & + , + & $139$ & 0 , - & $15$ \\
  Surprisal-GPT-2 + Energy (L5) & + , + & $\mathbf{468}$ & + , - & $121$ & 0 , + & $7$ \\
  Surprisal-GPT-2 + Energy (L6) & + , - & $337$ & + , + & $139$ & 0 , - & $15$ \\
  Entropy (L5) & + & $40$ & - & $44$ & + & $30$ \\
  Entropy (L6) & - & $8$ & - & $28$ & 0 & $1$ \\
  Surprisal + Entropy (L5) & + , + & $321$ & + , - & $112$ & 0 , + & $32$ \\
  Surprisal + Entropy (L6) & + , - & $320$ & + , - & $92$ & 0 , 0 & $2$ \\
  Surprisal-GPT-2 + Entropy (L5) & + , + & $330$ & + , - & $119$ & 0 , + & $\mathbf{32}$ \\
  Surprisal-GPT-2 + Entropy (L6) & + , - & $330$ & + , - & $105$ & 0 , 0 & $2$ \\
  Entropy (L5) + Energy (L5) & + , + & $211$ & - , - & $66$ & + , 0 & $30$ \\
  Entropy (L6) + Energy (L6) & - , - & $27$ & - , + & $125$ & 0 , - & $15$ \\
  Entropy-GPT2 (L12) & - & $2$ & - & $77$ & + & $20$ \\
  Surprisal + Entropy-GPT2 (L12) & + , - & $314$ & + , - & $127$ & 0 , + & $21$ \\
  Surprisal-GPT-2 + Entropy-GPT2 (L12) & + , - & $323$ & + , - & $133$ & 0 , + & $21$ \\
  Entropy-GPT2 (L11) + Energy (L5) & - , + & $254$ & - , - & $90$ & + , 0 & $25$ \\
  Entropy-GPT2 (L12) + Energy (L6) & - , - & $27$ & - , + & $121$ & + , - & $24$ \\
  Entropy-GPT2 (all layers) & - & $9$ & - & $149$ & + & $14$ \\
  Surprisal + Entropy-GPT2 (all layers) & + , - & $331$ & + , - & $195$ & 0 , + & $15$ \\
  Surprisal-GPT-2 + Entropy-GPT2 (all layers) & + , - & $340$ & + , - & $\mathbf{195}$ & 0 , + & $15$ \\
  Entropy-GPT2 (all layers) + Energy (L5) & - , + & $241$ & - , - & $167$ & + , + & $16$ \\
  Entropy-GPT2 (all layers) + Energy (L6) & - , - & $39$ & - , + & $173$ & + , - & $20$ \\
  \hline
\end{tabular}

  \caption{Mixed-effects model results for the Natural Stories (NSC) corpus and the self-paced reading (spr) and eye-tracking (et) portions of the UCL corpus filtered on single-token words. Format as in Table~\ref{tab:ns}. \label{tab:ns_single}}
\end{table*}

\end{document}